\documentclass[runningheads]{llncs}
\usepackage{graphicx}
\usepackage{booktabs}
\usepackage{placeins}
\usepackage{listings}
\usepackage{xcolor}

\newif\ifpublish
\publishtrue

\newcommand{\TensorQuant}{\textit{TensorQuant} }

\begin{document}
%
%
%
\title{Sparsity in Deep Neural Networks - An~Empirical Investigation with TensorQuant}
\titlerunning{Sparsity in Deep Neural Networks}
%

\ifpublish
\author{Dominik~Marek~Loroch\inst{1,2} \and 
	Franz-Josef~Pfreundt \inst{1} \and
	Norbert~Wehn \inst{2} \and
	Janis~Keuper\inst{1,3}
}
\authorrunning{D. M. Loroch et al.}

\institute{Fraunhofer ITWM, Germany\\
	\and TU Kaiserslautern, Germany\\
	\and Fraunhofer Center Machine Learning, Germany
}
\fi

\maketitle              

\begin{abstract}
Deep learning is finding its way into the embedded world with applications such as autonomous driving, smart sensors and augmented reality. However, the computation of deep neural networks is demanding in energy, compute power and memory. Various approaches have been investigated to reduce the necessary resources, one of which is to leverage the sparsity occurring in deep neural networks due to the high levels of redundancy in the network parameters. It has been shown that sparsity can be promoted specifically and the achieved sparsity can be very high. But in many cases the methods are evaluated on rather small topologies. It is not clear if the results transfer onto deeper topologies.\\
In this paper, the \TensorQuant toolbox has been extended to offer a platform to investigate sparsity, especially in deeper models. Several practical relevant topologies for varying classification problem sizes are investigated to show the differences in sparsity for activations, weights and gradients.
	
\end{abstract}
\FloatBarrier


\section{Introduction} \label{sec:intro}

For the past few years, deep learning had a high impact on machine learning. Many diverse applications have emerged in virtually all fields of research and everyday life. 
Initially being a high-performance computing problem, deep learning is finding its way into the mobile and embedded world with applications such as autonomous driving, smart sensors and augmented reality, just to name a few. There is a huge potential behind deep learning in the embedded world, where more and more of the heavy workload is moved to the device, known as edge computing.

However, the computation of deep neural networks (DNN) is very resource heavy in energy, compute power and memory, in both space and bandwidth. These problems have been circumvented by moving the data from the generating device at the edge to a centralized computation unit (i.e. cloud service). But as the number of devices and the demands for low latency increase, moving large amounts of data away from the device becomes infeasible.
The training of deep networks on distributed embedded systems is even more demanding, as it requires to send updates of all weights between the workers.

A key observation is that a large portion of the parameters in a neural network are redundant. If the operations and activations that are not necessary can be identified, this can make the calculation more efficient and save energy. It has been shown that models can be compressed by a factor of over 30 \cite{HanMao16Deep}.

\subsection{Related Work}
There have been several ideas on how to enable deep learning on the edge by removing redundancy.
A very well received approach is to use topologies which are specifically designed to be very lean and thus avoid redundancy by design
\cite{iandola2016squeezenet,HowZhu17MobileNet,ZhaZho17ShuffleNet}. 
This approach has a high popularity for mobile applications.

Instead of looking at the efficiency problem from an algorithmic side, the hardware can be adapted to be very efficient for computations required by DNNs.
The calculation of the operations in a layer can be moved to a co-processor. 


An option are FPGAs, which allow to design a specialized hardware architecture for DNNs, but at much less effort than building a computer chip from scratch. There are several examples of FPGA implementations dealing with redundancy in DNNs \cite{ZhuJia2017SparseNN,HanLiu16EIE,aimar2017nullhop,AndCav2016YodaNN,RybWeh17Hardware,ChaZai17Compiling}.
FPGAs consume little energy, therefore they are good candidates for embedded applications.

Redundancy reduction can also be seen in the context of distributed systems. Training DNNs on such systems is an active field of research \cite{YouGit17Large,KeuPfr16Distributed,KueKeu17Using}. A problem is the transfer of the weight updates in the form of gradients between the different nodes.
Redundancy appears in weight updates which have no effect on the convergence of the training. If that information can be prevented from being transferred, it can save bandwidth \cite{renggli2018sparcml,AjiFik2017Sparse,WanWan2018Gradient,rhu2018compressing}. It has been shown that compression ratios of up to 600 in memory size are possible \cite{LinHan17Deep}.

\subsection{Contribution}
All of the approaches above can be interpreted as a way to introduce and leverage sparsity in deep neural networks. Sparsity is the ratio of zero-value elements to all elements. This concept is applicable most directly to the weights in a neural network, but also reducing the number and shape of layers can be interpreted as a form of sparsity leveraging. 

The results for novel methods found in literature often use small topologies and simple datasets as a proof of concept. It is not clear if those results transfer well to bigger models, i.e. if these methods scale. For hardware accelerators, which have to rely on a certain amount of sparsity in order to be efficient, it is crucial to know if a certain topology is able to deliver unchanged results with a sparse representation of the data. There is a need for a methodology that can investigate the potential in sparsity of a model prior to the hardware implementation.

This paper extends the capabilities of \textit{TensorQuant}\footnote{www.tensor-quant.org} \cite{LorPfr17Tensorquant}, which is an open-source toolbox for \textit{TensorFlow} \cite{AbaBar16Tensorflow}. It can be used to investigate sparsity in custom topologies and datasets with very little changes to the original files describing the model. The contributions in this paper are:
\begin{itemize}
	\item Sparsity is studied in several convolutional neural network (CNN) topologies of varying sizes. The differences in the sparsity of the activations and weights during inference are investigated.
	\item The sparsity of the gradient during training is examined. This shows which level of accuracy can be expected for different gradient sparsity levels, if no additional methods are applied to guide the training process.
	\item \TensorQuant is extended and used to provide an easy way to access and manipulate the layers in a DNN for sparsity experiments. It offers an open platform to test and compare various methods which rely on tensor alteration, including sparsity.
\end{itemize}
This work puts methods which leverage sparsity into perspective, as it shows what level of sparsity can already emerge from using regular methods.

Chapter \ref{sec:methods} introduces the used terms and methods in this paper. It gives a brief overview of \textit{TensorQuant} and how it can help to investigate sparsity. In chapter \ref{sec:experiments} experiments are conducted, which show to which degree sparsity is emerging in CNNs, applying regular methods for training and inference.

\FloatBarrier

\section{Methods} \label{sec:methods}

A neural network layer is defined as
\begin{equation} \label{eq:layer}
\textrm{z}_l = f(\textrm{x}_l, \textrm{W}_l),
\end{equation}
where $\textrm{x}_l$ is the input, $\textrm{z}_l$ is the activation and $\textrm{W}_l$ is a set of weights in the layer $l$. $f$ is a non-linear function, called activation function. A neural network is trained by minimizing some loss function $\textrm{L}(\textrm{W})$, which can include terms for L1 and L2 regularization \cite{BotCur16Optimization}. The optimization step is
\begin{equation} \label{eq:gradient}
\textrm{w}_{t+1} = \textrm{w}_t - \lambda \frac{\partial\textrm{L}}{\partial\textrm{w}_t}
\end{equation}
for every weight $\textrm{w}$ in the neural network. $\frac{\partial\textrm{L}}{\partial\textrm{w}_t}$ is referred to as the gradient, and it is scaled with some learning rate $\lambda$ before applied to the weight as an update.

\subsection{Sparsity} \label{sec:sparsity}
Sparsity is defined as the ratio of zero-value elements to all elements.
The sparsity of a layer is
\begin{equation} \label{eq:layer_sparsity}
\textrm{s}_l=\frac{\left\vert {\left\{ {\textrm{w} \:|\: \textrm{w}=0, \textrm{w} \in \textrm{W}_l} \right\} } \right\vert } {\left\vert { \textrm{W}_l } \right\vert }.
\end{equation}
In a large model comprising of many layers it helps to group layers in a logical hierarchy, referred to as a block
\begin{equation}\label{eq:block}
\textrm{B}=\{l\:|\: \textrm{for some arbitrarily chosen } l\}.
\end{equation}
Then the sparsity of that block is the number of all zero weights divided by the number of weights belonging to that block
\begin{equation} \label{eq:block_sparsity}
\textrm{s}_b=\frac{\sum_{l \in \textrm{B}} \left\vert { \textrm{W}_l } \right\vert \textrm{s}_l } { \sum_{l \in \textrm{B}}\left\vert { \textrm{W}_l } \right\vert },
\end{equation}
with $\textrm{B}$ being the set of all layers belonging to the logical block. The total sparsity of a model can be calculated similarly, but summing over all layers in all blocks.

The gradients of the weights are grouped in the same way as the weights themselves and their sparsity is computed in the same manner.

The activation sparsity is different from the weights and gradients.
It always refers to the last activation of a block, without considering the other layers within the block as in equation (\ref{eq:block_sparsity}). It is defined similar to equation (\ref{eq:layer_sparsity}), but by counting over the activation values $\textrm{z}$ instead of weights $\textrm{w}$.

\subsection{Enforcing Sparsity} \label{sec:enforcing_sparsity}

When using a ReLU as the activation function, sparsity emerges in the activations to a high degree. For the weights and gradients, however, it is very unlikely that even one of their values is exactly zero. Applying equation (\ref{eq:layer_sparsity}) will always result in zero sparsity, as the filters are, in fact, dense. Therefore, it is necessary to enforce sparsity.
One method is to select a certain number of elements with the highest magnitude and set the other ones to zero \cite{SunRen17meProp}. Another way is to use a threshold for the magnitude and to set all the values below it to zero. The latter approach is used in this paper.\\

\subsection{TensorQuant}
\begin{figure}[htb]
	\centering
	\includegraphics[width=0.7\textwidth]{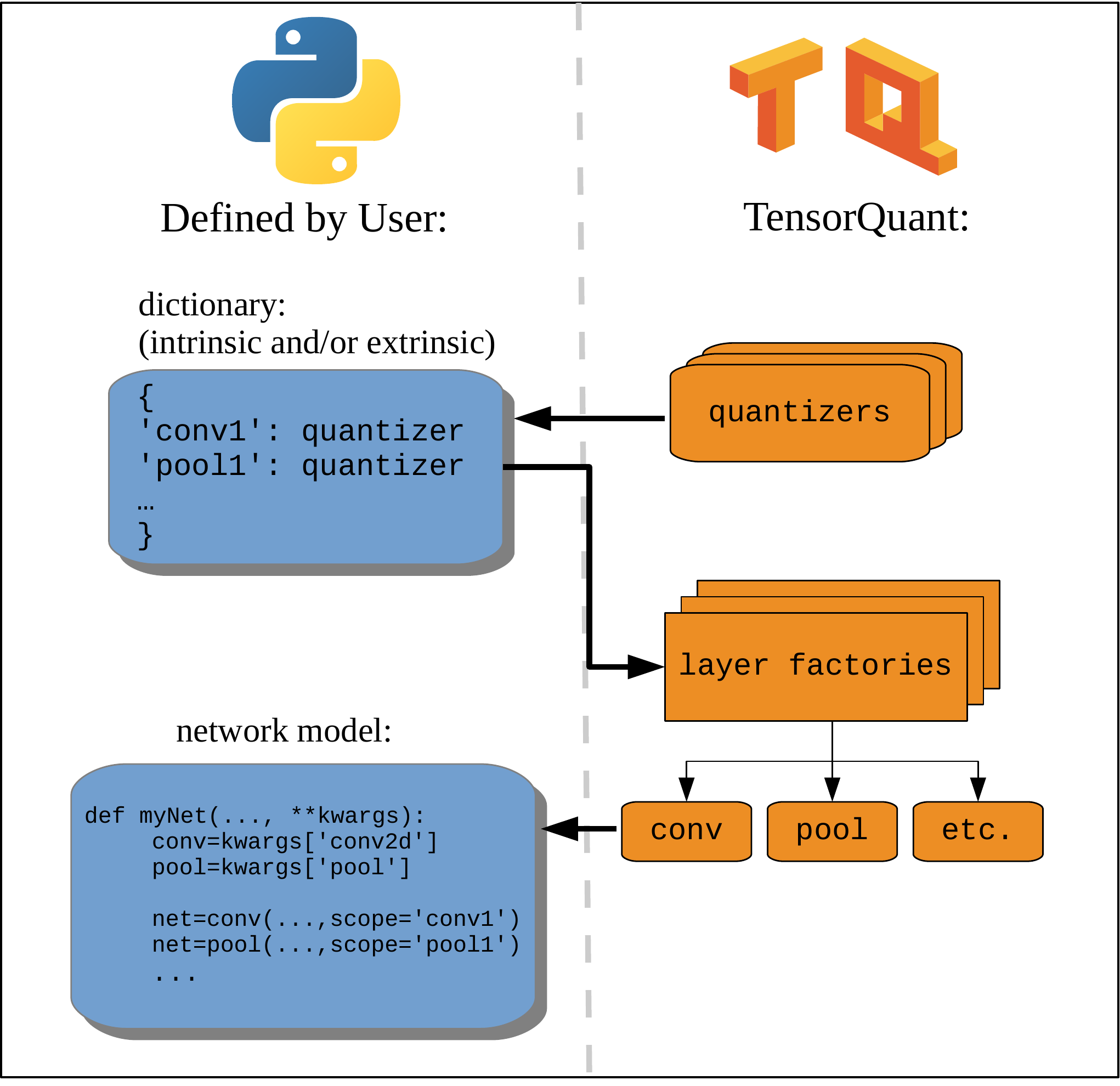}
	\caption{Overview of the TensorQuant workflow. The user provides a python dictionary, which maps variable scopes to quantizer objects. Minor changes need to be applied to the file describing the topology, so that \TensorQuant can loop in the quantizers at the desired locations.} \label{fig:TQ_OV}
\end{figure}
\TensorQuant is a toolbox for \textit{TensorFlow}, originally designed to investigate the effects of quantization on deep neural networks \cite{LorPfr17Tensorquant}. 
One of its distinct features is that it can manipulate the tensors in a network to a very deep level, without much changes to the files describing the model. Manipulation is performed by looping in additional operations at specific locations. \textit{TensorFlow} allows to introduce userdefined C++ kernels as additional tensor operators. 
In \textit{TensorQuant}, those operations are referred to as quantizers.
Thus, a quantizer or kernel can be designed which sets all entries to zero whose magnitude is below a certain threshold. By incorporating this operation into a model, the weights, activations and gradients can be sparsified. 

The lowest level where tensors can be manipulated in the context of quantization  is referred to as intrinsic quantization. Every layer is broken down to its tensor operations. The tensors passing from one operation to the next are quantized at every step in order to assure that the precision of the intermediate result never exceeds the one of the data format to be emulated. This way, \TensorQuant can emulate low-bitwidth operations, specifically in the convolution layers of CNNs.

Another location where tensor manipulation can be applied is just at the output of a layer. In the context of quantization, this is called extrinsic quantization, and it is the location where activation sparsification can be introduced. 
The weights can be manipulated just before they are passed to the operations, which allows to sparsify weights.
Also the gradients can be sparsified before they are being applied to the weights as updates.

\TensorQuant uses the \textit{TensorFlow} \textit{slim} framework to provide a variety of utility functions. \TensorQuant extends this framework by adding several additional functionalities which ease the access to the layers. See figure \ref{fig:TQ_OV} for an overview of the workflow. The layers in \textit{TensorFlow} are tagged with so called variable scopes. If a python dictionary is provided which maps these scopes to quantizers, \TensorQuant automatically applies tensor manipulation to the desired locations. Every layer and block can have their own quantizer. This automatism allows very deep topologies to be quantized easily with arbitrary granularity.

As for now, the \TensorQuant \textit{slim} utility collection is made for CNNs and classification tasks. However, \TensorQuant can be used on a much broader class of deep learning topologies.

Using the emulation capabilities of \TensorQuant comes at the cost of degrading the runtime. As for now, this renders training in combination with intrinsic quantization infeasible. Applying extrinsic quantization is much less problematic, so that training with sparsifying operators is not a problem.
\FloatBarrier

\section{Experiments} \label{sec:experiments}
This section investigates the effects of sparsity enforcement on some CNN classifiers. The choice of topologies reflects different difficulty levels. AlexNet \cite{KriSut12Imagenet} and ResNet~50 \cite{HeZha15Deep} are trained on the ILSVRC12 ImageNet \cite{DenDon09Imagenet} dataset, which is the most difficult task in this paper. ResNet~14  and CifarNet \cite{Kri12Learning} are trained on the CIFAR~100 and 10 \cite{AleVin09CIFAR} dataset, respectively. These two are considered to be medium and low level problems. Finally, the MNIST dataset is used to train LeNet \cite{lecun1998gradient}, which is considered to be a trivial problem.

AlexNet and CifarNet use dropout \cite{srivastava2014dropout}, whereas the Inception and ResNet topologies use batch normalization \cite{IofSze15Batch}. All topologies use ReLUs as activation functions. These special layers can have an additional impact on the sparsity, but which is not investigated in this paper.

The naming convention of the layers is as follows: "conv" refers to a single convolution layer, "logits" and "fc" to fully connected ones. In the ResNet topologies, a "block" is a logical block comprising of convolution layers with the same number of input and output filters. A "unit" comprises of a bottleneck layer \cite{HeZha15Deep}, which has three convolutions plus a shortcut connection.

\subsection{Sparsity of Activations and Weights}
If the weights of a DNN model are sparse, the required memory to store the model can be decreased. A high sparsity in activation values can decrease the computation time, even if the weights are not sparse. Therefore it is interesting to look at the sparsity of weights and activations.
Normally, a L2 regularizer is used during training. It is known that a L1 regularizer promotes sparsity in the weights, although it makes convergence to an optimum more difficult and therefore it is less often used. This section shows the different sparsity levels between a variety of CNN topologies, trained with L1 and L2 regularizers, respectively. The focus of this section is on the inference.

First, the network is trained without any sparsity enforcement with either L1 or L2 regularization.
As explained in section \ref{sec:enforcing_sparsity}, weights are not sparse as they are, so sparsity needs to be enforced, e.g. with thresholding. The objective here is to set as many weights to zero without retraining, so that the test accuracy is not changed. To obtain a high total sparsity, each layer or block has its own threshold. They are found with a grid search approach, by going through all layers or blocks iteratively. For each layer, the highest threshold is found which leaves the test accuracy unchanged. Meanwhile, the other layers are not sparsified. For the test accuracy, all thresholds for all layers are applied at once. The values for the test accuracies are stated in the captions of the respective figures, relative to the L2 test accuracy without sparsification.

Although activations could be sparsified with the same method, it has proven to be rather ineffective in our experience. The sparsity which comes from the ReLU activation functions is already high and further thresholding does not have much effect.\\

\begin{table}
	\centering
	\caption{LeNet weight and activation sparsity after training with L1 and L2 regularizer. The relative test accuracies are L1 99.0\%  and L2 99.8\%.}
	\label{tab:lenet_sparsity}
	\setlength{\tabcolsep}{1em}
	\begin{tabular}{lcccc}
	\toprule
	Layer & L2 weights & L1 weights & L2 activations & L1 activations\\ 
	\midrule
		conv1  & 0.142 & 0.289 & 0.717 & 0.513\\ 
		conv2  & 0.491 & 0.505 & 0.528 & 0.662\\ 
		fc3  & 0.258 & 0.502 & 0.661 & 0.593\\ 
		fc4  & 0.258 & 0.504 & 0.000 & 0.000\\ 
	\bottomrule
	\end{tabular}
\end{table}

Table \ref{tab:lenet_sparsity} shows the weight and activation sparsities for LeNet trained with L1 and L2 regularization, respectively. L1 increases the sparsity of weights in every layer, especially in the last two fully connected layers. The activations change in sparsity as well, though there is no general trend. The last layer is the classification output, so it is no surprise that the sparsity is zero. Notice that in LeNet the test accuracy is higher with L1 regularization than for L2.

\begin{table}
	\centering
	\caption{CifarNet weight and activation sparsity after training with L1 and L2 regularizer. The relative test accuracies are L1 101.8\% and L2 98.4\%.}
	\label{tab:cifarnet_sparsity}
	\setlength{\tabcolsep}{1em}
	\begin{tabular}{lcccc}
	\toprule
	Layer & L2 weights & L1 weights & L2 activations & L1 activations\\ 
	\midrule
		conv1  & 0.064 & 0.117 & 0.683 & 0.665\\ 
		conv2  & 0.226 & 0.754 & 0.854 & 0.837\\ 
		fc3  & 0.164 & 0.623 & 0.781 & 0.797\\ 
		fc4  & 0.563 & 0.501 & 0.649 & 0.458\\ 
		logits  & 0.862 & 0.120 & 0.000 & 0.000\\ 
	\bottomrule
	\end{tabular}
\end{table}

CifarNet in table \ref{tab:cifarnet_sparsity} shows a more dramatic increase in weight sparsity in some of its layers. But surprisingly, the sparsity in the last layer dropped a lot. The activation sparsities, however, are mostly unchanged between L1 and L2.

\begin{table}
	\centering
	\caption{ResNet~14 weight and activation sparsity after training with L1 and L2 regularizer. The relative test accuracies are L1 93.2\% and L2 99.6\%.}
	\label{tab:resnet14_sparsity}
	\setlength{\tabcolsep}{1em}
	\begin{tabular}{lcccc}
	\toprule
	Layer & L2 weights & L1 weights & L2 activations & L1 activations\\ 
	\midrule
		conv1  & 0.051 & 0.382 & 0.285 & 0.273\\ 
		block1/unit\_1  & 0.188 & 0.390 & 0.428 & 0.407\\ 
		block1/unit\_2  & 0.074 & 0.501 & 0.274 & 0.257\\ 
		block2/unit\_1  & 0.058 & 0.313 & 0.673 & 0.699\\ 
		block2/unit\_2  & 0.056 & 0.349 & 0.766 & 0.745\\ 
		logits  & 0.028 & 0.233 & 0.000 & 0.000\\ 
	\bottomrule
	\end{tabular}
\end{table}

The ResNet~14 topology in table \ref{tab:resnet14_sparsity} exhibits a very low weight sparsity for L2 and rises only moderately with L1. The activation sparsity does not change between L1 and L2, as it was the case with CifarNet.

\begin{table}
	\caption{AlexNet weight and activation sparsity after training with L2 regularizer. The relative test accuracy is L2 98.0\%.} \label{tab:alexnet_sparsity}
	\centering
	\setlength{\tabcolsep}{1em}
	\begin{tabular}{lcc}
	\toprule
	Layer & L2 weights & L2 activations\\ 
	\midrule
		conv1  & 0.161 & 0.604\\ 
		conv2  & 0.177 & 0.804\\ 
		conv3  & 0.177 & 0.825\\ 
		conv4  & 0.406 & 0.863\\ 
		conv5  & 0.219 & 0.920\\ 
		fc6  & 0.524 & 0.817\\ 
		fc7  & 0.730 & 0.838\\ 
		fc8  & 0.474 & 0.000\\ 
	\bottomrule
	\end{tabular}
\end{table}
\ifx
\begin{table}
	\caption{AlexNet weight and activation sparsity after training with L1 and L2 regularizer. The relative test accuracies are L2 98.0\% and L1 19.2\% } \label{tab:alexnet_sparsity-2}
	\centering
	\setlength{\tabcolsep}{1em}
	\begin{tabular}{lcccc}
		\toprule
		Layer & L2 weights & L1 weights & L2 activations & L1 activations\\ 
		\midrule
		conv1  & 0.161 & 0.805 & 0.604 & 0.843\\ 
		conv2  & 0.177 & 0.942 & 0.804 & 0.762\\ 
		conv3  & 0.177 & 0.981 & 0.825 & 0.934\\ 
		conv4  & 0.406 & 0.986 & 0.863 & 0.977\\ 
		conv5  & 0.219 & 0.997 & 0.920 & 0.983\\ 
		fc6  & 0.524 & 0.999 & 0.817 & 0.078\\ 
		fc7  & 0.730 & 0.998 & 0.838 & 0.031\\ 
		fc8  & 0.474 & 0.981 & 0.000 & 0.000\\ 
		\bottomrule
	\end{tabular}
\end{table}
\fi
Training AlexNet with a L1 regularizer is difficult, and even when incorporating a mixed L1-L2 regularization, the results remain poor, so only the results for L2 regularization are shown.
The L2 weight sparsity for AlexNet in table \ref{tab:alexnet_sparsity} is low for most of the convolution layers, but high for the fully connected ones. The activation sparsity is very high in all layers.

\begin{table} [htbp]
	\caption{ResNet~50 weight and activation sparsity after training with L2 regularizer. The relative test accuracy is L2 99.6\% }
	\setlength{\tabcolsep}{1em}
	\centering
	\label{tab:resnet50_sparsity}
	\begin{tabular}{lcc}
		\toprule
		Layer & L2 weights & L2 activations \\ 
		\midrule
		conv1  & 0.073 & 0.301\\ 
		block1/unit\_1  & 0.378 & 0.359\\ 
		block1/unit\_2  & 0.369 & 0.236\\ 
		block1/unit\_3  & 0.359 & 0.214\\ 
		block2/unit\_1  & 0.122 & 0.493\\ 
		block2/unit\_2  & 0.489 & 0.343\\ 
		block2/unit\_3  & 0.126 & 0.304\\ 
		block2/unit\_4  & 0.125 & 0.268\\ 
		block3/unit\_1  & 0.177 & 0.542\\ 
		block3/unit\_2  & 0.344 & 0.417\\ 
		block3/unit\_3  & 0.178 & 0.369\\ 
		block3/unit\_4  & 0.370 & 0.336\\ 
		block3/unit\_5  & 0.391 & 0.306\\ 
		block3/unit\_6  & 0.210 & 0.270\\ 
		block4/unit\_1  & 0.242 & 0.735\\ 
		block4/unit\_2  & 0.510 & 0.737\\ 
		block4/unit\_3  & 0.265  & 0.824\\ 
		logits  & 0.127 & 0.000\\ 
		\bottomrule
	\end{tabular}
\end{table}
\ifx
\begin{table}
	\caption{ResNet~50 weight and activation sparsity after training with L1 and L2 regularizer. The relative test accuracies are L2 99.6\% and L1 54.3\% }
	\setlength{\tabcolsep}{1em}
	\label{tab:resnet50_sparsity2}
	\begin{tabular}{lcccc}
	\toprule
	Layer & L2 weights & L1 weights & L2 activations & L1 activations\\ 
	\midrule
		conv1  & 0.073 & 0.803 & 0.301 & 0.496\\ 
		block1/unit\_1  & 0.378 & 0.925 & 0.359 & 0.398\\ 
		block1/unit\_2  & 0.369 & 0.941 & 0.236 & 0.276\\ 
		block1/unit\_3  & 0.359 & 0.859 & 0.214 & 0.256\\ 
		block2/unit\_1  & 0.122 & 0.934 & 0.493 & 0.552\\ 
		block2/unit\_2  & 0.489 & 0.901 & 0.343 & 0.441\\ 
		block2/unit\_3  & 0.126 & 0.970 & 0.304 & 0.399\\ 
		block2/unit\_4  & 0.125 & 0.936 & 0.268 & 0.372\\ 
		block3/unit\_1  & 0.177 & 0.957 & 0.542 & 0.618\\ 
		block3/unit\_2  & 0.344 & 0.993 & 0.417 & 0.612\\ 
		block3/unit\_3  & 0.178 & 0.969 & 0.369 & 0.635\\ 
		block3/unit\_4  & 0.370 & 0.986 & 0.336 & 0.597\\ 
		block3/unit\_5  & 0.391 & 0.990 & 0.306 & 0.506\\ 
		block3/unit\_6  & 0.210 & 0.998 & 0.270 & 0.525\\ 
		block4/unit\_1  & 0.242 & 0.947 & 0.735 & 0.933\\ 
		block4/unit\_2  & 0.510 & 0.998 & 0.737 & 0.940\\ 
		block4/unit\_3  & 0.265 & 0.993 & 0.824 & 0.965\\ 
		logits  & 0.127 & 0.973 & 0.000 & 0.000\\ 
	\bottomrule
	\end{tabular}
\end{table}
\fi
ResNet~50 also trains poorly with the L1 regularizer. For L2, table \ref{tab:resnet50_sparsity} shows that the weight sparsity is not high for most of the layers. Activation sparsity is rather low. The "unit 1" layers in every block have the highest activation sparsity, except for the layers in block~4, where it is very high in all units.

\begin{table}[htbp]
	\caption{Overview of total weight sparsity after training with L1 and L2 regularizer.}
	\centering
	\setlength{\tabcolsep}{1em}
	\label{tab:L1-L2-sparsity_OV}
	\begin{tabular}{lcc}
	\toprule
	Layer & L2 & L1\\ 
	\midrule
		LeNet  & 0.262 & 0.502\\
		CifarNet  & 0.185 & 0.624\\  
		ResNet~14  & 0.062 & 0.326\\ 
		AlexNet   & 0.573 & - \\
		ResNet~50   & 0.291 & - \\
		
	\bottomrule
	\end{tabular}
\end{table}

Table \ref{tab:L1-L2-sparsity_OV} gives an overview of the total weight sparsities for L1 and L2 regularization. It shows that for simpler problems, it is easy to achieve high sparsity even with simple regularization methods.
In all topologies, the weight sparsity is lower than the one for activations, which agrees with observations made in other work \cite{WenWu2016Learning,LiuPoo2018Efficient,ZhuJia2017SparseNN,rhu2018compressing}. 
Identifying layers with sparse activations contains valuable information for model parallelism. They are good locations to cut the topology into subgraphs, which can be put on seperate nodes in a distributed system. This allows to minimize the amount of data which needs to be transferred. For instance, the "unit 1" layers of each block in ResNet~50 would be good separation points.

\subsection{Sparsity of Gradients during Training}

When training on a distributed system, the sparsity in the gradients can help to reduce the amount of data which needs to be transferred to compute an update. So in this section, the sparsity of the gradient is investigated during training. Similar to the weights, sparsity needs to be enforced. In a single training run, the same threshold is applied to all gradients in every step. L2 regularization is used during training.

\begin{figure}[htbp]
	\centering
	\includegraphics[width=0.6\textwidth]{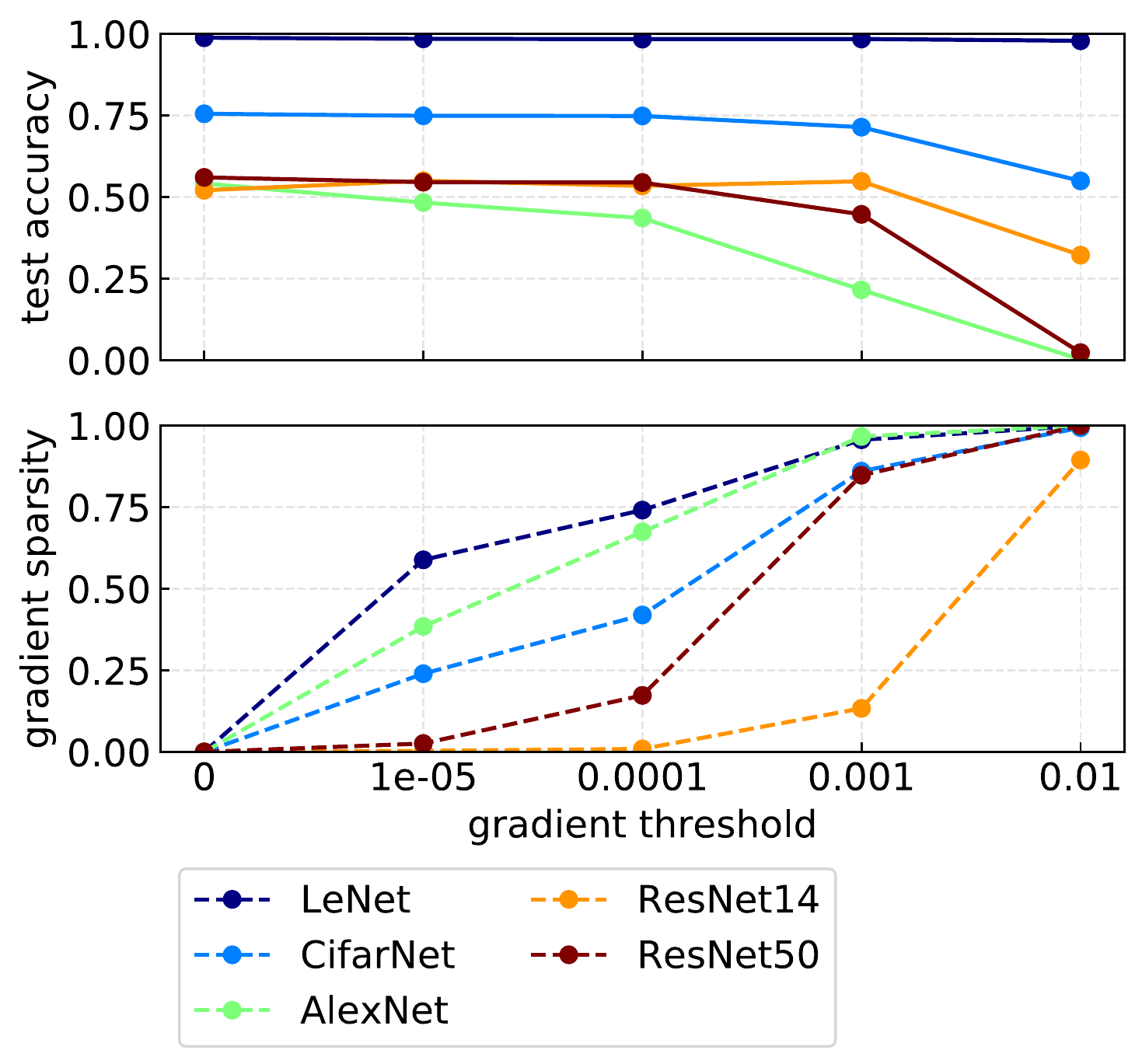}
	\caption{Comparison of absolute test accuracy versus the applied gradient threshold during training.} \label{fig:grad_vs_acc}
\end{figure}
Figure \ref{fig:grad_vs_acc} shows the final test accuracy and the total gradient sparsity towards the end of the training versus the applied gradient threshold. All training runs have the same number of iterations as the baseline. A threshold of zero indicates the baseline. AlexNet is more susceptible for sparsified gradients than the ResNet topologies. For ResNet14, the sparse gradients have a regularizing effect, so the test accuracy increases above the baseline if the threshold is not too high. Such a regularizing effect has also been observed in other work \cite{SunRen17meProp}. CifarNet is mostly unchanged similarly to ResNet~14, but does not show the same regularizing effect. The LeNet topology is almost unchanged for the investigated thresholds.

The gradient sparsity can be almost 100\,\% for LeNet and the model can still learn fine. This indicates that MNIST on LeNet is a rather trivial problem. 
AlexNet shows a steady decline in test accuracy with an increasing threshold. CifarNet and the two ResNet architectures have a jump in gradient sparsity, but where the test accuracy does not change much. ResNet~50 can achieve 80\,\% baseline accuracy at a gradient sparsity of 85\,\%. This suggests that there is a sweet spot for the gradient threshold, which allows for very high sparsities in those topologies.

\begin{figure}[htbp]
	\centering
	\includegraphics[width=0.6\textwidth]{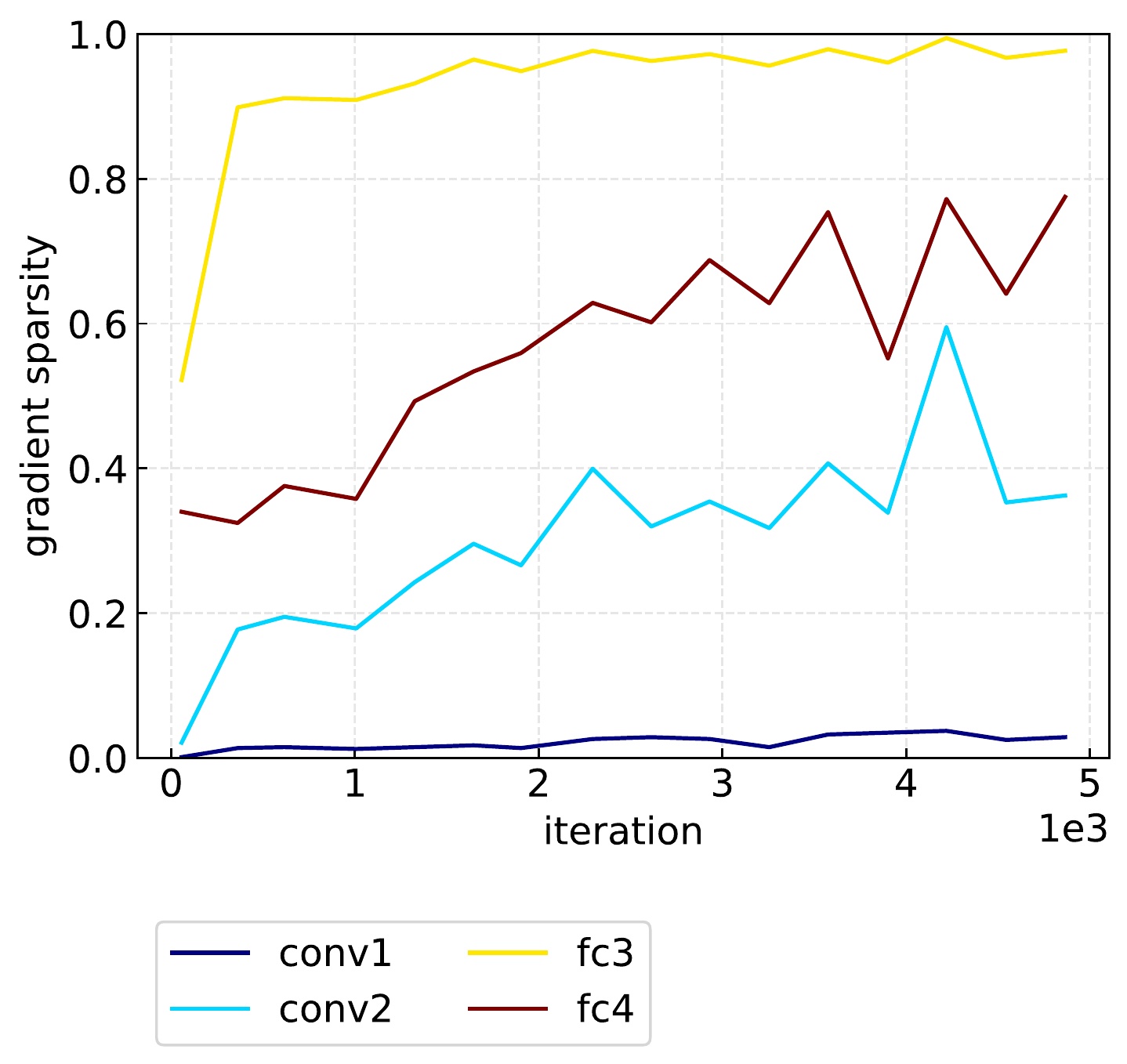}
	\caption{LeNet gradient sparsity of different layers during training. The achieved test accuracy is 99\,\% of the baseline. The gradient threshold is $10^{-3}$.} \label{fig:lenet_grad}
\end{figure}

\begin{figure}[htbp]
	\centering
	\includegraphics[width=0.6\textwidth]{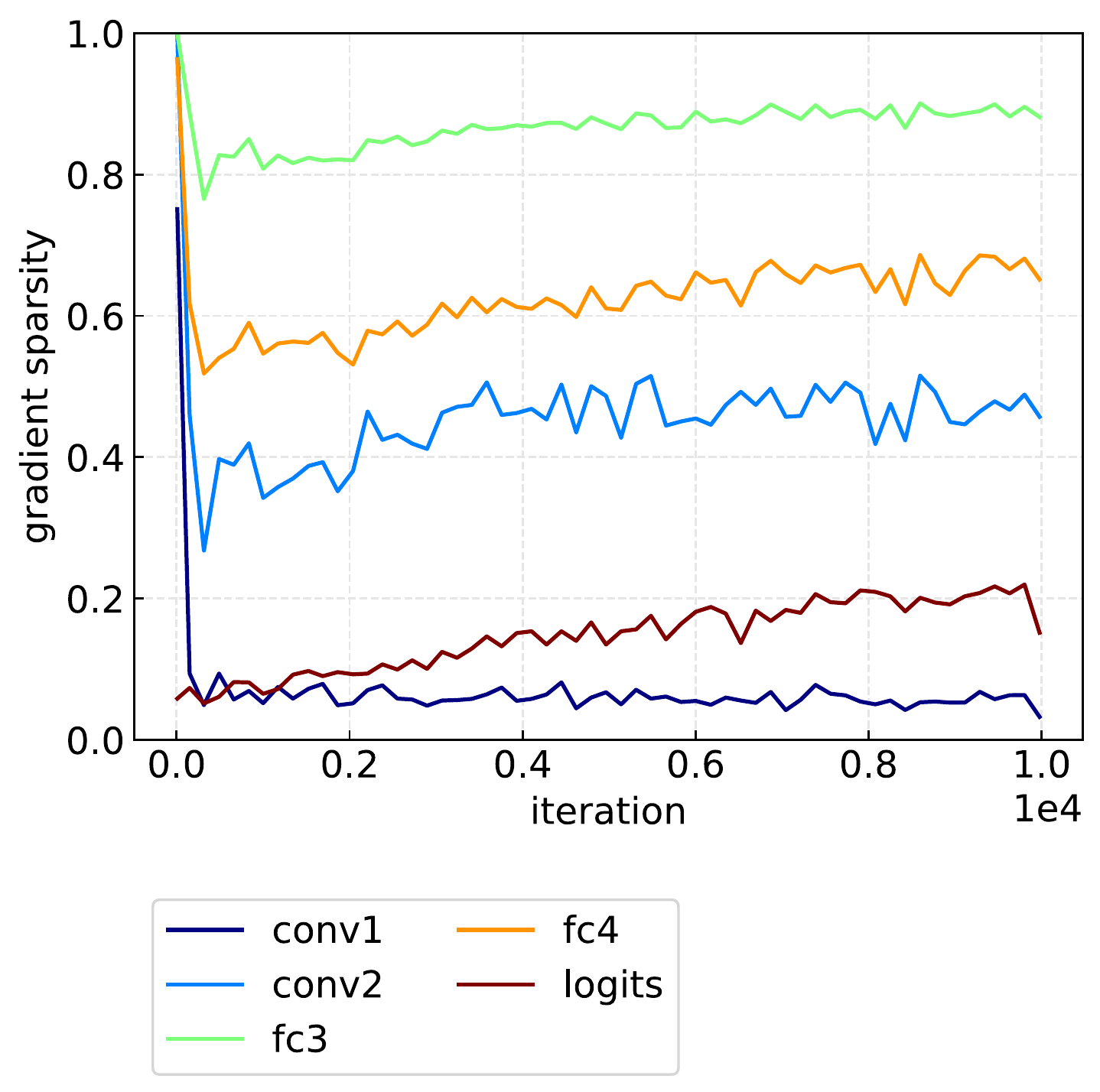}
	\caption{CifarNet gradient sparsity of different layers during training. The achieved test accuracy is 97\,\% of the baseline. The gradient threshold is $10^{-3}$.} \label{fig:cifarnet_grad}
\end{figure}

\begin{figure}[htbp]
	\centering
	\includegraphics[width=0.6\textwidth]{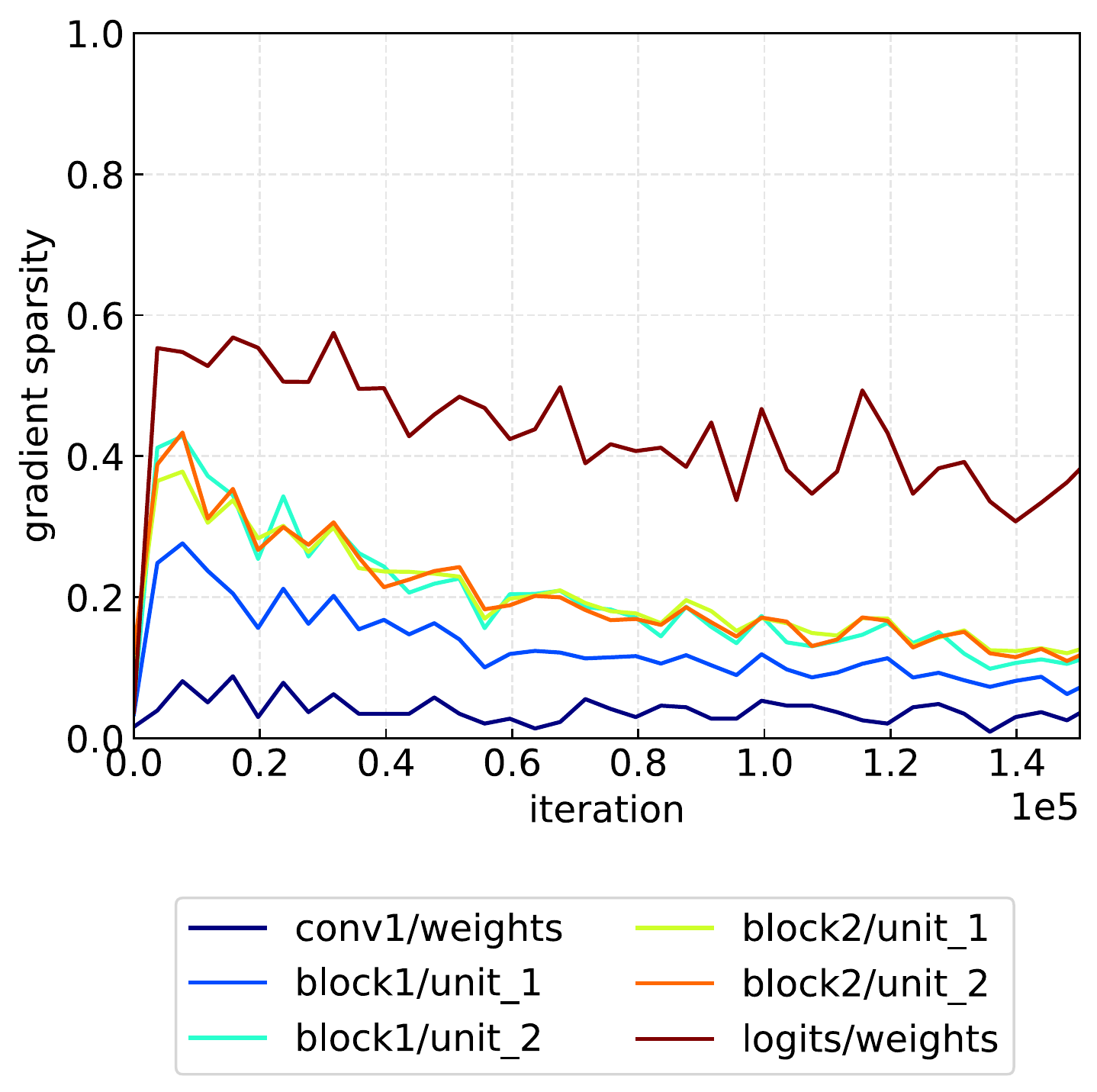}
	\caption{ResNet~14 gradient sparsity of different layers during training. The achieved test accuracy is 105\,\% of the baseline. The gradient threshold is $10^{-3}$.} \label{fig:resnet14_grad}
\end{figure}

\begin{figure}[htbp]
	\centering
	\includegraphics[width=0.6\textwidth]{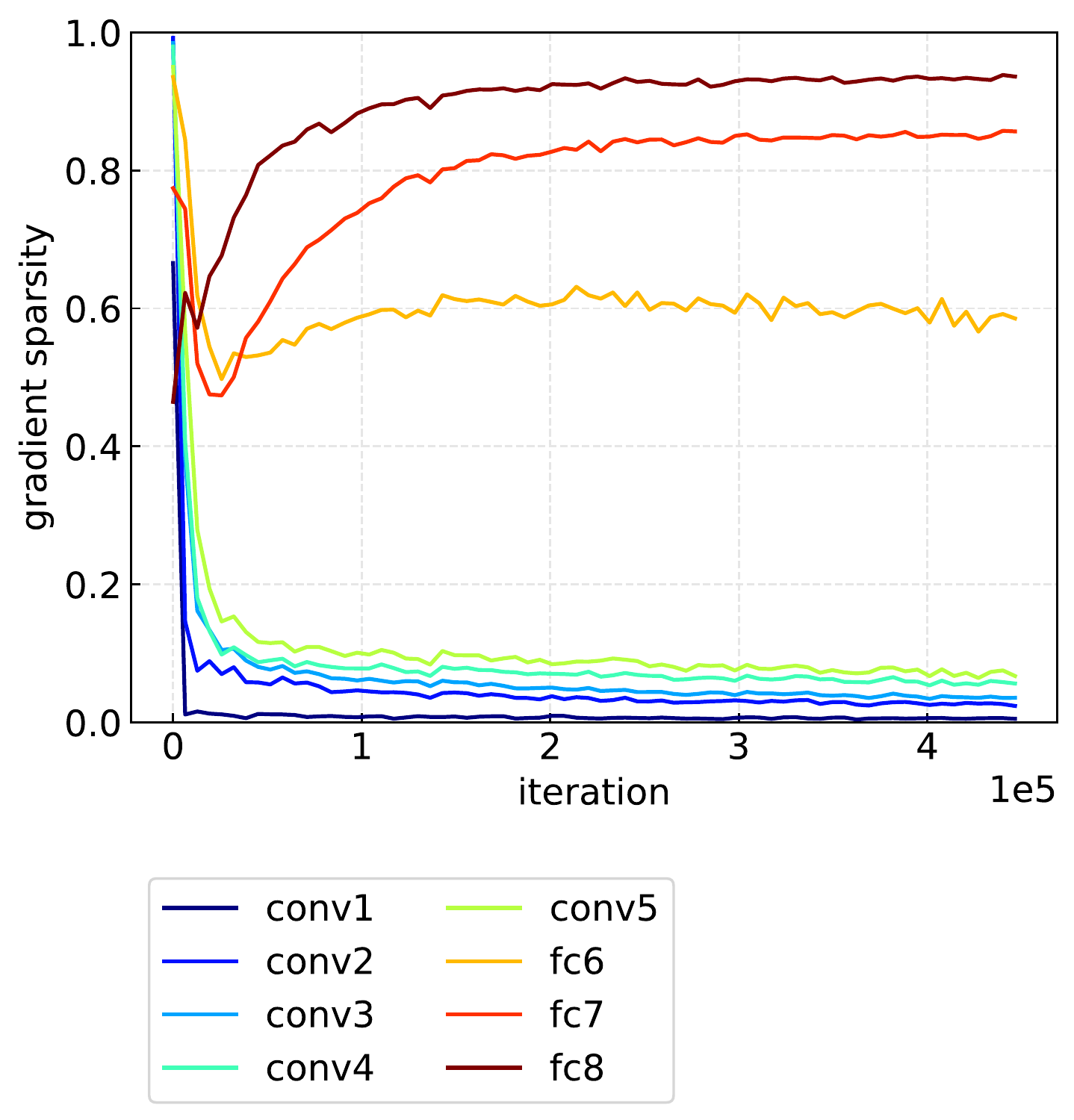}
	\caption{AlexNet gradient sparsity of different layers during training. The achieved test accuracy is 80\,\% of the baseline. The gradient threshold is $10^{-4}$.} \label{fig:alexnet_grad}
\end{figure}

\begin{figure}[htb]
	\centering
	\includegraphics[width=0.6\textwidth]{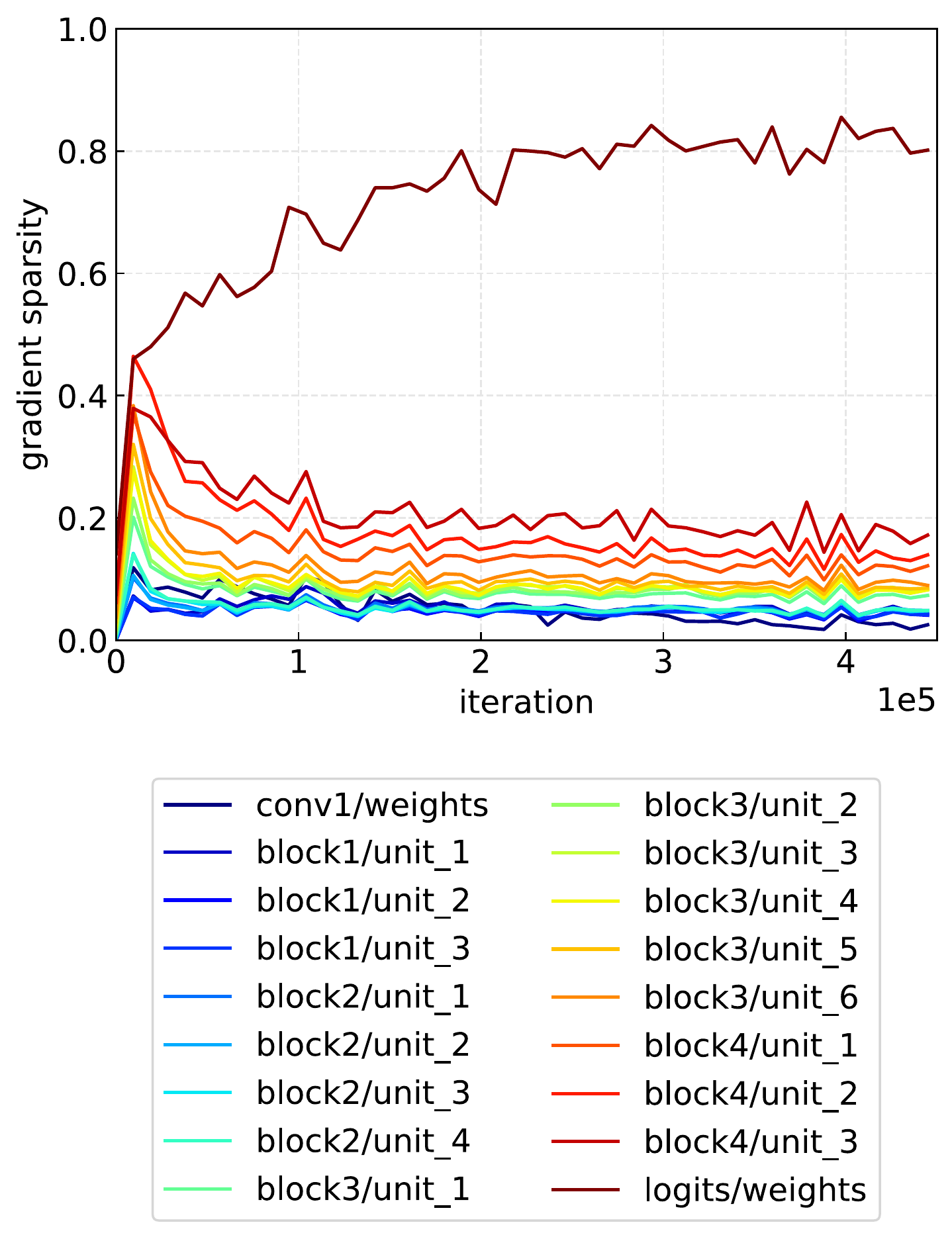}
	\caption{ResNet~50 gradient sparsity of different layers during training. The achieved test accuracy is 96\,\% of the baseline. The gradient threshold is $10^{-4}$.} \label{fig:resnet50_grad}
\end{figure}

Figures \ref{fig:lenet_grad} to \ref{fig:resnet50_grad} show how the gradient sparsity evolves during training for individual layers or blocks. The weights are initialized with a gaussian distribution. AlexNet, LeNet and CifarNet are trained with a batch size of 128, and 32 for the ResNet topologies. The gradient thresholds are chosen in such a way that the final test accuracy is close to the baseline accuracy, but also that there is some visible gradient sparsity. The gradient thresholds and achieved accuracies are stated in the captions of the figures.

The layerwise gradient sparsity for LeNet is shown in figure \ref{fig:lenet_grad}. The gap between the first layer and fc3 is striking. This suggests that conv1 holds the most information, whereas fc3 is redundant.

The evolution of the gradient sparsities in CifarNet differ somewhat from LeNet, even though the topologies are very similar. There is a very high peak in sparsity at a very early phase of the training, except for the logits layer. Each layer seems to converge towards a certain sparsity level, where the fc3 layer, which is in the middle of the topology, has the highest sparsity.

Resnet~14 in figure \ref{fig:resnet14_grad} comprises almost entirely of convolution layers, only the last layer is fully connected. All gradient sparsities increase rapidly in the first epoch, then they show a decreasing trend. The logits layer has a higher sparsity than all the other layers. Different to the two topologies before, the convolution layers exhibit a similar, low sparsity. The decreasing trend in gradient sparsity seems to contradict the fact that the gradient is becoming flatter the closer the weights converge to an optimum. However, the decrease in gradient sparsity only means that the number of non-zero elements is increasing, it does not imply anything about the magnitude of the gradient itself. A possible explanation is that the gradient is pointing more equally into multiple dimensions when it gets closer to an optimum than at the beginning of the training.

For AlexNet in figure \ref{fig:alexnet_grad}, the fully connected layers show a more pronounced gap to the convolution layers than the topologies before. The fully connected layers also have a more unique behavior. There is an initial decline in sparsity in the beginning, followed by an increase after a few epochs. Then, the gradient sparsities converge toward different levels. The seemingly same sparsity level for the convolution layers is a artifact introduced by the chosen threshold. A higher value would spread out the sparsity levels, which is not shown here.

ResNet~50 in figure \ref{fig:resnet50_grad} is similar to ResNet~14, but there is a bigger gap between the sparsity level of the last, fully connected layer and the other convolutional ones. Also, the gradient sparsity for the last fully connected layer goes up instead of down.\\

The figures above give a good overview of the relative behavior of the gradient sparsities of different layers. The absolute values of the sparsities are less meaningful, since the thresholds are chosen rather arbitrarily (figure \ref{fig:grad_vs_acc} is a better reference in that regard). The most striking result is that the gradient sparsity of convolution layers decreases in the more complex topologies, which seemingly contradicts the fact that the gradient becomes flatter.

\FloatBarrier

\section{Conclusion} \label{sec:conclusion}

Experiments have been conducted on a selection of CNN topologies, showing sparsity for weights, activations and gradients under changing problem size.
Although all of them are CNN classifiers, there are differences in where and to which degree sparsity emerges, especially in the gradients during training.
The training of LeNet on MNIST has been shown to be a trivial problem, which requires almost no gradient information to be trained close to 100\,\% test accuracy. Therefore, results obtained from a less complex topology cannot be transfered to deeper networks. It is necessary to investigate sparsity for each topology and sparsifying method on their own in order to get meaningful information about sparsity.

In many cases there already is a moderate degree of sparsity in the regularly trained versions of the models. The application of additional methods to promote sparsity can increase the levels beyond the results shown here, but this paper serves as a reference point for what can be expected from the baseline model.

Our results back the idea of implementing sparse arithmetics on embedded devices, since the redundancy in form of sparsity can be leveraged through special hardware architectures.
\TensorQuant can help in the investigation of sparsity in deep neural networks by identifying where sparsity emerges to a high degree.
The information obtained from this can guide the design of sparse arithmetics hardware accelerators.
\TensorQuant is open-source and freely available on \textit{GitHub}\footnote{www.tensor-quant.org}.
\FloatBarrier

%
%
%
\bibliographystyle{ieeetr}
\bibliography{TensorQuant}

\begin{thebibliography}{10}

\bibitem{HanMao16Deep}
S.~Han, H.~Mao, and W.~J. Dally, ``Deep compression: Compressing deep neural
  networks with pruning, trained quantization and huffman coding,'' 2015.

\bibitem{iandola2016squeezenet}
F.~N. Iandola, S.~Han, M.~W. Moskewicz, K.~Ashraf, W.~J. Dally, and K.~Keutzer,
  ``Squeezenet: Alexnet-level accuracy with 50x fewer parameters and <0.5mb
  model size,'' 2016.

\bibitem{HowZhu17MobileNet}
A.~G. Howard, M.~Zhu, B.~Chen, D.~Kalenichenko, W.~Wang, T.~Weyand,
  M.~Andreetto, and H.~Adam, ``Mobilenets: Efficient convolutional neural
  networks for mobile vision applications,'' 2017.

\bibitem{ZhaZho17ShuffleNet}
X.~Zhang, X.~Zhou, M.~Lin, and J.~Sun, ``Shufflenet: An extremely efficient
  convolutional neural network for mobile devices,'' 2017.

\bibitem{ZhuJia2017SparseNN}
J.~Zhu, J.~Jiang, X.~Chen, and C.-Y. Tsui, ``Sparsenn: An energy-efficient
  neural network accelerator exploiting input and output sparsity.,'' {\em
  CoRR}, vol.~abs/1711.01263, 2017.

\bibitem{HanLiu16EIE}
S.~Han, X.~Liu, H.~Mao, J.~Pu, A.~Pedram, M.~A. Horowitz, and W.~J. Dally,
  ``Eie: Efficient inference engine on compressed deep neural network.,'' in
  {\em ISCA}, pp.~243--254, IEEE Computer Society, 2016.

\bibitem{aimar2017nullhop}
A.~Aimar, H.~Mostafa, E.~Calabrese, A.~Rios-Navarro, R.~Tapiador-Morales, I.-A.
  Lungu, M.~B. Milde, F.~Corradi, A.~Linares-Barranco, S.-C. Liu, and
  T.~Delbruck, ``Nullhop: A flexible convolutional neural network accelerator
  based on sparse representations of feature maps,'' 2017.

\bibitem{AndCav2016YodaNN}
R.~Andri, L.~Cavigelli, D.~Rossi, and L.~Benini, ``Yodann: An architecture for
  ultra-low power binary-weight cnn acceleration,'' 2016.

\bibitem{RybWeh17Hardware}
V.~Rybalkin, N.~Wehn, M.~R. Yousefi, and D.~Stricker, ``Hardware architecture
  of bidirectional long short-term memory neural network for optical character
  recognition,'' in {\em Proceedings of the Conference on Design, Automation \&
  Test in Europe}, pp.~1394--1399, European Design and Automation Association,
  2017.

\bibitem{ChaZai17Compiling}
A.~X.~M. Chang, A.~Zaidy, V.~Gokhale, and E.~Culurciello, ``Compiling deep
  learning models for custom hardware accelerators,'' 2017.

\bibitem{YouGit17Large}
Y.~You, I.~Gitman, and B.~Ginsburg, ``Large batch training of convolutional
  networks,'' 2017.

\bibitem{KeuPfr16Distributed}
J.~Keuper and F.-J. Pfreundt, ``Distributed training of deep neural networks:
  Theoretical and practical limits of parallel scalability,'' 2016.

\bibitem{KueKeu17Using}
M.~Kuehn, J.~Keuper, and F.-J. Pfreundt, ``Using gpi-2 for distributed memory
  paralleliziation of the caffe toolbox to speed up deep neural network
  training,'' 2017.

\bibitem{renggli2018sparcml}
C.~Renggli, D.~Alistarh, and T.~Hoefler, ``Sparcml: High-performance sparse
  communication for machine learning,'' 2018.

\bibitem{AjiFik2017Sparse}
A.~F. Aji and K.~Heafield, ``Sparse communication for distributed gradient
  descent,'' 2017.

\bibitem{WanWan2018Gradient}
J.~Wangni, J.~Wang, J.~Liu, and T.~Zhang, ``Gradient sparsification for
  communication-efficient distributed optimization,'' 2017.

\bibitem{rhu2018compressing}
M.~Rhu, M.~O'Connor, N.~Chatterjee, J.~Pool, Y.~Kwon, and S.~W. Keckler,
  ``Compressing dma engine: Leveraging activation sparsity for training deep
  neural networks,'' in {\em High Performance Computer Architecture (HPCA),
  2018 IEEE International Symposium on}, pp.~78--91, IEEE, 2018.

\bibitem{LinHan17Deep}
Y.~Lin, S.~Han, H.~Mao, Y.~Wang, and W.~J. Dally, ``Deep gradient compression:
  Reducing the communication bandwidth for distributed training,'' 2017.

\bibitem{LorPfr17Tensorquant}
D.~M. Loroch, F.-J. Pfreundt, N.~Wehn, and J.~Keuper, ``Tensorquant: A
  simulation toolbox for deep neural network quantization.,'' in {\em
  MLHPC@SC}, pp.~1:1--1:8, ACM, 2017.

\bibitem{AbaBar16Tensorflow}
M.~Abadi, P.~Barham, J.~Chen, Z.~Chen, A.~Davis, J.~Dean, M.~Devin,
  S.~Ghemawat, G.~Irving, M.~Isard, {\em et~al.}, ``Tensorflow: A system for
  large-scale machine learning.,'' in {\em OSDI}, vol.~16, pp.~265--283, 2016.

\bibitem{BotCur16Optimization}
L.~Bottou, F.~E. Curtis, and J.~Nocedal, ``Optimization methods for large-scale
  machine learning,'' 2016.

\bibitem{SunRen17meProp}
X.~Sun, X.~Ren, S.~Ma, and H.~Wang, ``meprop: Sparsified back propagation for
  accelerated deep learning with reduced overfitting.,'' {\em CoRR},
  vol.~abs/1706.06197, 2017.

\bibitem{KriSut12Imagenet}
A.~Krizhevsky, I.~Sutskever, and G.~E. Hinton, ``Imagenet classification with
  deep convolutional neural networks,'' in {\em Advances in neural information
  processing systems}, pp.~1097--1105, 2012.

\bibitem{HeZha15Deep}
K.~He, X.~Zhang, S.~Ren, and J.~Sun, ``Deep residual learning for image
  recognition,'' 2015.

\bibitem{DenDon09Imagenet}
J.~Deng, W.~Dong, R.~Socher, L.-J. Li, K.~Li, and L.~Fei-Fei, ``Imagenet: A
  large-scale hierarchical image database,'' in {\em Computer Vision and
  Pattern Recognition, 2009. CVPR 2009. IEEE Conference on}, pp.~248--255,
  IEEE, 2009.

\bibitem{Kri12Learning}
A.~Krizhevsky, ``Learning multiple layers of features from tiny images,'' 05
  2012.

\bibitem{AleVin09CIFAR}
A.~Krizhevsky, V.~Nair, and G.~Hinton, ``Cifar-100 (canadian institute for
  advanced research),'' 2009.

\bibitem{lecun1998gradient}
Y.~LeCun, L.~Bottou, Y.~Bengio, and P.~Haffner, ``Gradient-based learning
  applied to document recognition,'' {\em Proceedings of the IEEE}, vol.~86,
  no.~11, pp.~2278--2324, 1998.

\bibitem{srivastava2014dropout}
N.~Srivastava, G.~E. Hinton, A.~Krizhevsky, I.~Sutskever, and R.~Salakhutdinov,
  ``Dropout: a simple way to prevent neural networks from overfitting,'' {\em
  Journal of machine learning research}, vol.~15, no.~1, pp.~1929--1958, 2014.

\bibitem{IofSze15Batch}
S.~Ioffe and C.~Szegedy, ``Batch normalization: Accelerating deep network
  training by reducing internal covariate shift.,'' {\em CoRR},
  vol.~abs/1502.03167, 2015.

\bibitem{WenWu2016Learning}
W.~Wen, C.~Wu, Y.~Wang, Y.~Chen, and H.~Li, ``Learning structured sparsity in
  deep neural networks,'' 2016.

\bibitem{LiuPoo2018Efficient}
X.~Liu, J.~Pool, S.~Han, and W.~J. Dally, ``Efficient sparse-winograd
  convolutional neural networks.,'' {\em CoRR}, vol.~abs/1802.06367, 2018.

\end{thebibliography}

\end{document}